\documentclass[conference,a4paper]{APSIPA2021}
\usepackage{amsmath}
\usepackage{amssymb}
\usepackage{graphicx}
\usepackage{multirow}
\usepackage{mathtools}

\usepackage[backend=biber,style=ieee,]{biblatex}
\usepackage{geometry}
\usepackage{fancyhdr}

\fancypagestyle{firststyle}{
  \fancyhf{}
  \fancyhead[C]{2023 Asia Pacific Signal and Information Processing Association Annual Summit and Conference (APSIPA ASC)}
}

\begin{document}

\title{CPIPS: Learning to Preserve Perceptual Distances in End-to-End Image Compression}

\author{
\authorblockN{
Chen-Hsiu Huang and Ja-Ling Wu \\
CMLab, CSIE, National Taiwan University, Taiwan \\
E-mail: \{chenhsiu48,wjl\}@cmlab.csie.ntu.edu.tw
}


}

\maketitle
\thispagestyle{firststyle}

\begin{abstract}
Lossy image coding standards such as JPEG and MPEG have successfully achieved high compression rates for human consumption of multimedia data. However, with the increasing prevalence of IoT devices, drones, and self-driving cars, machines rather than humans are processing a greater portion of captured visual content. Consequently, it is crucial to pursue an efficient compressed representation that caters not only to human vision but also to image processing and machine vision tasks. Drawing inspiration from the efficient coding hypothesis in biological systems and the modeling of the sensory cortex in neural science, we repurpose the compressed latent representation to prioritize semantic relevance while preserving perceptual distance. Our proposed method, Compressed Perceptual Image Patch Similarity (CPIPS), can be derived at a minimal cost from a learned neural codec and computed significantly faster than DNN-based perceptual metrics such as LPIPS and DISTS.
\end{abstract}


\section{Introduction}

The concept of \textit{efficient coding} \cite{attneave1954some,barlow1961possible} in early biological sensory processing systems hypothesized that the internal representation of images in the human visual system is optimized to encode the visual information it processes efficiently. In other words, the brain effectively compresses visual information.

The field of neural science has made discoveries regarding modeling neural single-unit and population responses in higher visual cortical areas using goal-driven hierarchical convolutional neural networks (HCNNs) \cite{yamins2016using}. The sensory cortex's fundamental framework models the visual system through encoding, the process by which stimuli are transformed into patterns of neural activity, and decoding, the process by which neural activity generates behavior. In their work \cite{yamins2016using}, HCNNs have successfully described the mapping of stimuli to measured neural responses in the brain.

In recent years, the rapid advancement of deep neural network techniques has significantly improved computer vision tasks \cite{krizhevsky2017imagenet,ronneberger2015u,bochkovskiy2020yolov4} and image processing tasks \cite{zhang2017beyond,dong2014learning,yu2018generative}. Neural compression \cite{yang2022introduction}, an end-to-end learned image compression method \cite{balle2016end,balle2018variational,minnen2018joint,minnen2020channel,cheng2020learned,chen2021end,duan2023lossy}, has also gained significant attention and has been shown to outperform traditional expert-designed image codecs. Traditionally, most image processing algorithms cannot be directly applied to hand-crafted image codecs like JPEG \cite{wallace1992jpeg}. As a result, the first step before further image processing or analysis is typically decompressing the image into raw pixels. With the evolution of neural compression, there is a growing trend to apply CNN-based methods directly to the compressed latent space \cite{huang2023neuralrep,testolina2021towards,le2021learned,duan2023unified}, leveraging the advantages of joint compression-accuracy optimization \cite{le2021learned} and eliminating the need for decompression. Consequently, international standards such as JPEG AI \cite{ascenso2021white} and MPEG VCM (Video Coding for Machines) \cite{evidence2020vcm} have been initiated to bridge data compression and computer vision, catering to both human and machine vision needs.

\begin{figure}[!t]
\centering
\includegraphics[width=1\columnwidth]{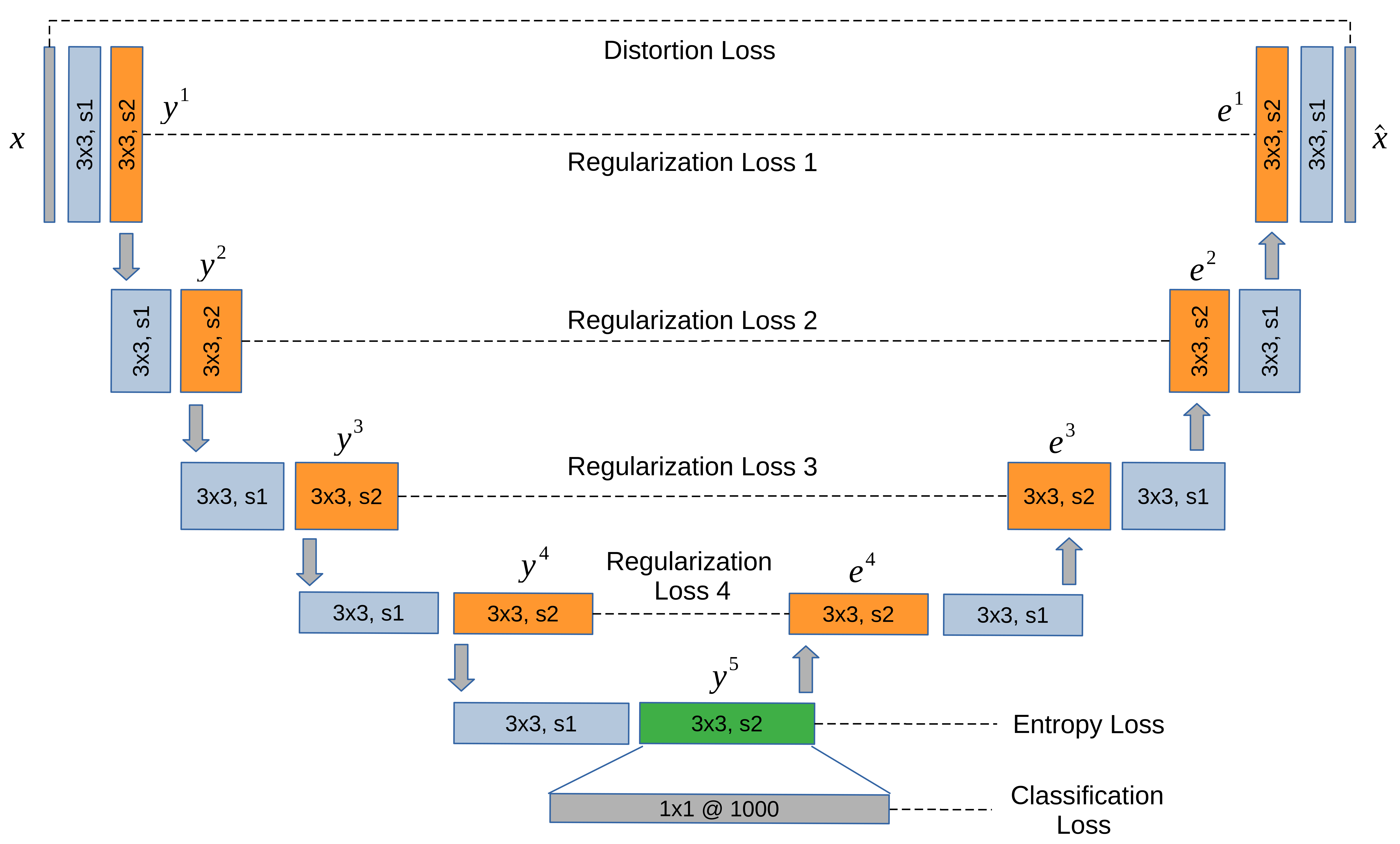}
\caption{The architecture we proposed for conducting perceptual distance preserving image compression. The innermost green convolution output $y^5$ represents the compressed latents to be further entropy-coded. The orange layer outputs $y^l$ and $e^l$, trained with the image classification task, contain semantic features that preserve perceptual differences in human vision.}
\label{fig:arch}
\end{figure}

Drawing inspiration from sensory cortex modeling \cite{yamins2016using} and the efficient coding hypothesis employed in information-theoretic perceptual quality metrics \cite{bhardwaj2020unsupervised}, we aim to develop an end-to-end learned image compression method jointly trained with the ImageNet classification task as the goal-driven HCNN. Fig. \ref{fig:arch} illustrates our proposed architecture, which resembles a UNet network. The compressed latent representations and the intermediate decoder output layers are mapped to a semantic space that preserves the perceptual distance between two different images. We name our method the \textit{Compressed} Perceptual Image Patch Similarity (CPIPS), which utilizes the entropy-coded bitstream and intermediate decoder output to measure the perceptual distance between images. In the context of Coding for Machines, the compressed image bitstream transmitted by IoT devices can be readily utilized by machines to assess perceptual distortions resulting from image operations.

Our contributions can be summarized as follows:

\begin{itemize}
\item We demonstrate the utilization of a goal-driven HCNN as an auxiliary task to map the latent space of the end-to-end learned image compression method to a space with semantic meaning.
\item We provide guidance and insights on designing the network architecture when a high-level computer vision task is jointly trained with a variational autoencoder network.
\item The proposed perceptual metric, CPIPS, is lightweight compared to other CNN-based perceptual metrics, such as LPIPS \cite{zhang2018perceptual} and DISTS \cite{ding2020image}. Computing CPIPS is significantly faster than LPIPS, with an acceleration of approximately 50 times.
\end{itemize}

\section{Related Works}

\subsection{Learned Image Compression}

The field of learned image compression has witnessed significant advancements with the introduction of convolutional neural networks. Several approaches have been proposed in the literature, starting with Ballé et al. \cite{balle2018variational} that surpassed traditional codecs like JPEG \cite{wallace1992jpeg} and JPEG 2000 \cite{adams2001jpeg} in terms of PSNR and SSIM metrics. Minnen et al. \cite{minnen2018joint} further improved coding efficiency by employing a joint autoregressive and hierarchical prior model, surpassing the performance of the HEVC \cite{lainema2016hevc} codec. More recently, Cheng et al. \cite{cheng2020learned} developed techniques that achieved comparable performance to the latest coding standard VVC \cite{ohm2018versatile}. Several comprehensive survey and introduction papers \cite{ma2019image,mishra2022deep,yang2022introduction} have summarized these advancements in end-to-end learned compression.

Currently, there are two remaining challenges \cite{keynote2023dcc} in this field: computational complexity and subjective image quality. The neural compressor employs high-capacity networks to end-to-end model data dependency in exchange for better bitrate-distortion (BD) efficiency. The channel-conditional method proposed by Minnen et al. \cite{minnen2020channel} achieves performance close to VVC but at the cost of high computational complexity (600K FLOPS/pixel). Regarding image quality, Valenzise et al. \cite{valenzise2018quality} conducted subjective tests on DNN-based methods and observed that these methods produce artifacts that are difficult to evaluate using traditional metrics like PSNR. They concluded that PSNR is inadequate for evaluating DNN-based methods. Upenik et al. \cite{upenik2021large} benchmarked a set of DNN-based image codecs using a crowdsourcing-based subjective quality evaluation procedure with Differential Mean Opinion Scores (DMOS). Their results demonstrate that learning-based approaches can achieve promising bitrate-DMOS performance compared to HEVC. However, despite their superior subjective scores, these DNN-based image codecs are optimized with pixel difference-based distortion functions.

\subsection{Perceptual Quality Metrics}

The evaluation of image codec quality traditionally relies on full-reference image quality assessment (FR-IQA) metrics, which measure the similarity between the reconstructed image and the original image as perceived by human observers. In addition, to mean square error (MSE) or PSNR, various FR-IQA metrics, such as SSIM variants \cite{wang2004image,wang2003multiscale}, PIM \cite{bhardwaj2020unsupervised}, and DISTS \cite{ding2020image}, have been proposed to predict subjective image quality judgments. Johnson et al. \cite{johnson2016perceptual} proposed using the feature vector distance from the VGG network \cite{simonyan2014very} as a perceptual loss for image transformation tasks based on the hypothesis that the same image features used for image classification are also helpful for other tasks.

Zhang et al. \cite{zhang2018perceptual} introduced the BAPPS dataset, which includes a large-scale collection of human judgments on image pairs, and trained the Learned Perceptual Image Patch Similarity (LPIPS) metric. LPIPS was found to be more aligned with human judgments than traditional quality metrics such as L2, PSNR, and SSIM. Ding et al. \cite{ding2021comparison} conducted an interesting study to evaluate whether DNN-based quality metrics can be used as objectives for optimizing image processing algorithms. Developing effective perceptual quality metrics for image tasks remains a challenging problem.

\subsection{Coding for Machines}

Lossy image coding standards such as JPEG and MPEG have primarily focused on achieving high compression rates for human consumption of multimedia data. However, with the rise of IoT devices, drones, and self-driving cars, there is a growing need for efficient compressed representations that cater not only to human vision but also to image processing and machine vision tasks. Techniques such as image data hiding \cite{huang2023neuralrep}, image denoising \cite{testolina2021towards}, and image super-resolution \cite{upenik2021towards} have been developed to operate directly on neural compressed latent spaces.

Le et al. \cite{le2021learned} proposed an inference-time content-adaptive fine-tuning scheme that optimizes the latent representation to improve compression efficiency for machine consumption. Duan et al. \cite{duan2023unified} employed transfer learning to perform semantic inference directly from quantized latent features in the deep compressed domain without pixel reconstruction. Choi et al. \cite{choi2022scalable} introduced scalable image coding frameworks based on well-developed neural compressors, achieving up to 80\% bitrate savings for machine vision tasks.

\section{Proposed Methods}

To enable joint training of the image compression network and an image classification task, one has to design a suitable network architecture that can be shared between a variational encoder network $\mathcal{G}_e$ and a DNN feature extraction network $\mathcal{F}$. We leverage the successful UNet \cite{ronneberger2015u} and VGG \cite{simonyan2014very} networks and propose a Left-UNet. Our Left-UNet consists of $L=5$ downsampling convolution layers, each with two convolution blocks. As shown in Fig. \ref{fig:arch}, the first orange block from the top-left represents the intermediate encoder output feature $y^1$ from the second convolution block of the first layer, denoted as \textbf{conv\_1\_2}. In Fig. \ref{fig:arch}, the innermost latent vector $y^5$, colored green, is outputted from \textbf{conv\_5\_2}. This vector is subject to quantization, resulting in an approximation $\hat{y}^5=\mathsf{Q}(y^5)$, which is then entropy-coded.

\subsection{Image Classification} \label{sec:imagenet}

\begin{figure}[!t]
\centering
\includegraphics[width=1\columnwidth]{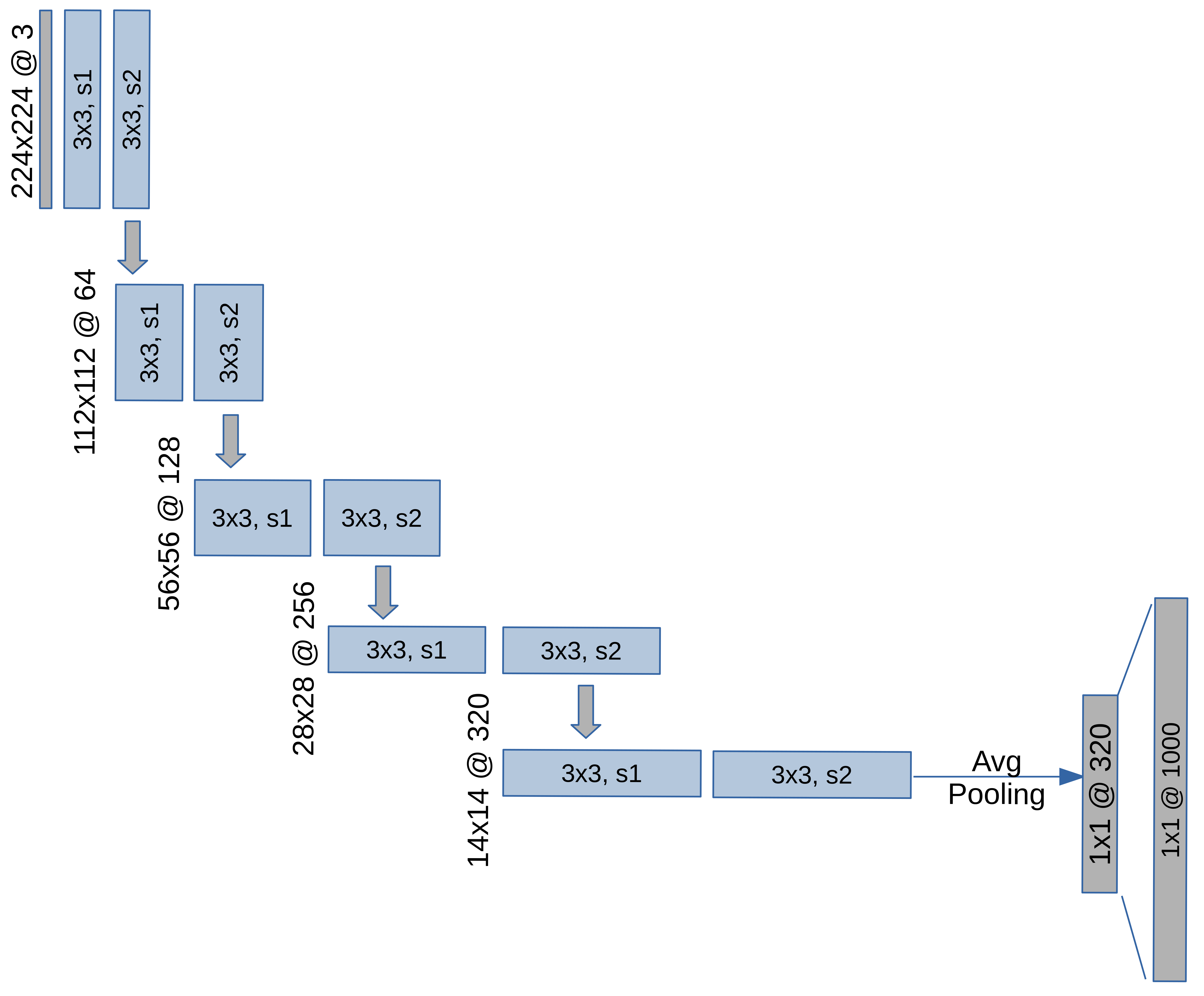}
\caption{The Left-UNet feature extraction network $\mathcal{F}$ and the classifier we proposed for the image classification task.}
\label{fig:leftunet}
\end{figure}

We illustrate the Left-UNet architecture in Fig. \ref{fig:leftunet} and provide details in Table \ref{tab:encoder}. The feature extraction network $\mathcal{F}$ for the image classification task uses the parameterized ReLU as the activation function for all layers, while the encoder network $\mathcal{G}_e$ employs Generalized Divisive Normalization (GDN) at the end of each downsampling layer. GDN, proposed by Ballé et al. \cite{balle2016end_2}, is inspired by modeling neurons in biological visual systems and has been proven effective in Gaussianizing image densities for a superior rate-distortion trade-off.

The extracted image features are then average pooled and connected to a linear layer with 1,000 neurons to optimize the classification loss $\mathcal{L}_C$ using cross-entropy:

\begin{equation}
\mathcal{L}_C = - \sum_i t_i \log(\mathcal{F}(x)_i)
\end{equation}

It is known that a high-capacity neural network trained for a high-level vision task implicitly learns to reason about relevant semantics \cite{johnson2016perceptual}. Our goal is not to solve the classification problem directly. Instead, we aim to design a moderately-sized network that can learn semantic features without significantly increasing the encoder-decoder complexity.

\subsection{Image Compression Network}

A typical learned neural codec consists of an encoder-decoder pair, a quantization module, and an entropy coder. Given an input image $x \in \mathcal{X}$, the neural encoder $\mathcal{G}_e$ transforms $x$ into a latent representation $y=\mathcal{G}_e(x)$, which is later quantized to a discrete-valued vector $\hat{y}$. The discrete probability distribution $P_{\hat{y}}$ is estimated using a neural network and then encoded into a bitstream using an entropy coder. The \textit{rate} of this discrete code, $\mathsf{R}$, is lower-bounded by the entropy of the discrete probability distribution $H(P_{\hat{y}})$. On the decoder side, we decode $\hat{y}$ from the bitstream and reconstruct the image $\hat{x}=\mathcal{G}_d(\hat{y})$ using the neural decoder. The \textit{distortion}, $\mathsf{D}$, is measured by a perceptual metric $d(x,\hat{x})$. Overall, we optimize the network parameters for a weighted sum of the rate and distortion, $\mathsf{R} + \lambda \mathsf{D}$, over a set of images.

Table \ref{tab:decoder} illustrates the decoder network $\mathcal{G}_d$, which is designed to complement the encoder. In the generic neural codec concept, the innermost latent vector $\hat{y}^5$ is equivalent to the discrete-valued vector $\hat{y}$. During image reconstruction, the intermediate output vectors $e^l$ from each upsampling layer \textbf{conv\_l\_2} play a crucial role because they represent learned multi-scale semantic layers, which are equivalent to the feature layers of a VGG-16 network.

\begin{table}[!ht]
\centering
\caption{Left-UNet architecture for $\mathcal{G}_e$ and $\mathcal{F}$}
\begin{tabular}{lrrrrc}
\hline
Layer & Kernel & Stride & In & Out & Output \\ \hline
conv\_1\_1 & 3 & 1 & 3 & 32 &  \\
PReLU &  &  &  &  &  \\
conv\_1\_2 & 3 & 2 & 32 & 32 &  \\
PReLU or GDN &  &  &  &  & $y^1$ \\
conv\_2\_1 & 3 & 1 & 32 & 64 &  \\
PReLU &  &  &  &  &  \\
conv\_2\_2 & 3 & 2 & 64 & 64 &  \\
PReLU or GDN &  &  &  & & $y^2$ \\
conv\_3\_1 & 3 & 1 & 64 & 128 &  \\
PReLU &  &  &  &  &  \\
conv\_3\_2 & 3 & 2 & 128 & 128 &  \\
PReLU or GDN &  &  &  & & $y^3$ \\
conv\_4\_1 & 3 & 1 & 128 & 256 &  \\
PReLU &  &  &  &  &  \\
conv\_4\_2 & 3 & 2 & 256 & 256 &  \\
PReLU or GDN &  &  &  & & $y^4$ \\
conv\_5\_1 & 3 & 1 & 256 & 320 &  \\
PReLU &  &  &  &  &  \\
conv\_5\_2 & 3 & 2 & 320 & 320 & $y^5$ \\
\hline
\end{tabular} \label{tab:encoder}
\end{table}

\begin{table}[!ht]
\centering
\caption{Decoder network architecture $\mathcal{G}_d$}
\begin{tabular}{lrrrrc}
\hline
Layer & Kernel & Stride& In & Out & Output \\ \hline
deconv\_5\_1 & 3 & 2 & 320 & 320 &  \\
PReLU &  &  &  &  &  \\
conv\_5\_2 & 3 & 1 & 320 & 256 & \\ 
GDN &  &  & &  & $e^4$ \\
deconv\_4\_1 & 3 & 2 & 256 & 256 &  \\
PReLU &  &  &  &  &  \\
conv\_4\_2 & 3 & 1 & 256 & 128 & \\ 
GDN &  &  & &  & $e^3$ \\
deconv\_3\_1 & 3 & 2 & 128 & 128 &  \\
PReLU &  &  &  &  &  \\
conv\_3\_2 & 3 & 1 & 128 & 64 & \\ 
GDN &  &  & &  & $e^2$ \\
deconv\_2\_1 & 3 & 2 & 64 & 64 &  \\
PReLU &  &  &  &  &  \\
conv\_2\_2 & 3 & 1 & 64 & 32 & \\ 
GDN &  &  & &  & $e^1$ \\
deconv\_1\_1 & 3 & 2 & 32 & 32 &  \\
PReLU &  &  &  &  &  \\
conv\_1\_2 & 3 & 1 & 32 & 3 & $\hat{x}$ \\ 
\hline
\end{tabular} \label{tab:decoder}
\end{table}

Like \cite{balle2018variational}, we employ kernel density estimation with a neural network to obtain the probability distribution $P_{\hat{y}}$. The rate loss $\mathsf{R}$ is computed as follows:

\begin{equation}
\mathsf{R} = -\mathbb{E}[\log_2 P_{\hat{y}}]
\end{equation}

In our experiments, we utilize the MSE as the distortion function. However, alternative quality metrics such as SSIM variants \cite{wang2004image, wang2003multiscale} can be employed to fit perceptual quality better. The distortion loss $\mathsf{D}$ is defined as:

\begin{equation}
\mathsf{D} = \mathbb{E}[d(x,\hat{x})] = \mathbb{E}||x-\hat{x} ||_2^2
\end{equation}

\subsection{Joint Compression-Classification Learning}

Although the intermediate convolution output features are seldom used in most machine learning tasks, these features, which are tuned to be predictive of essential structures, exhibit a high correlation with human perceptual similarity \cite{zhang2018perceptual}. However, storing intermediate latent features in the context of data compression becomes impractical if the final bottleneck layer contains sufficient information for the decoder to reconstruct the image. Another approach to mitigate storage waste is to reduce the number of downsampling layers. However, modeling the sensory cortex in the visual system \cite{yamins2016using} requires at least five layers of feature extraction to generate neural responses, a finding that our experiments validate as well. Consequently, we utilize the intermediate output $e^l$ from the decoder as a proxy for multi-scale semantic features and apply a regularizer to constrain the decoder. Specifically, we employ the $l_1$ distance to define our regularization loss:

\begin{equation}
\mathcal{L}_R = \sum_{l=1}^4 ||e^l-y^l||_1
\end{equation}

To initialize the Left-UNet encoder $\mathcal{G}_e$ and an auxiliary classifier, we utilize pre-trained semantic features from the image classification task mentioned in Section \ref{sec:imagenet}. Subsequently, we train an end-to-end image compression network using the overall loss function:

\begin{equation}
\mathcal{L} = \mathsf{R} + \lambda \mathsf{D} + \alpha \mathcal{L}_C + \beta \mathcal{L}_R
\end{equation}

The hyper-parameter $\lambda$ represents the rate-distortion trade-off, which can be adjusted according to the desired image quality factor $Q$. We set $\alpha=0.3$ and $\beta=1.0$ for our experiments.

Through joint compression-classification training, the weights of the Left-UNet encoder are initially initialized with pre-trained semantic features. Subsequently, the gradient descent optimizer updates the encoder-decoder weights to analyze and synthesize the image while improving classification accuracy.

\subsection{Compressed Perceptual Image Patch Similarity}

\begin{figure}[!t]
\centering
\includegraphics[width=1\columnwidth]{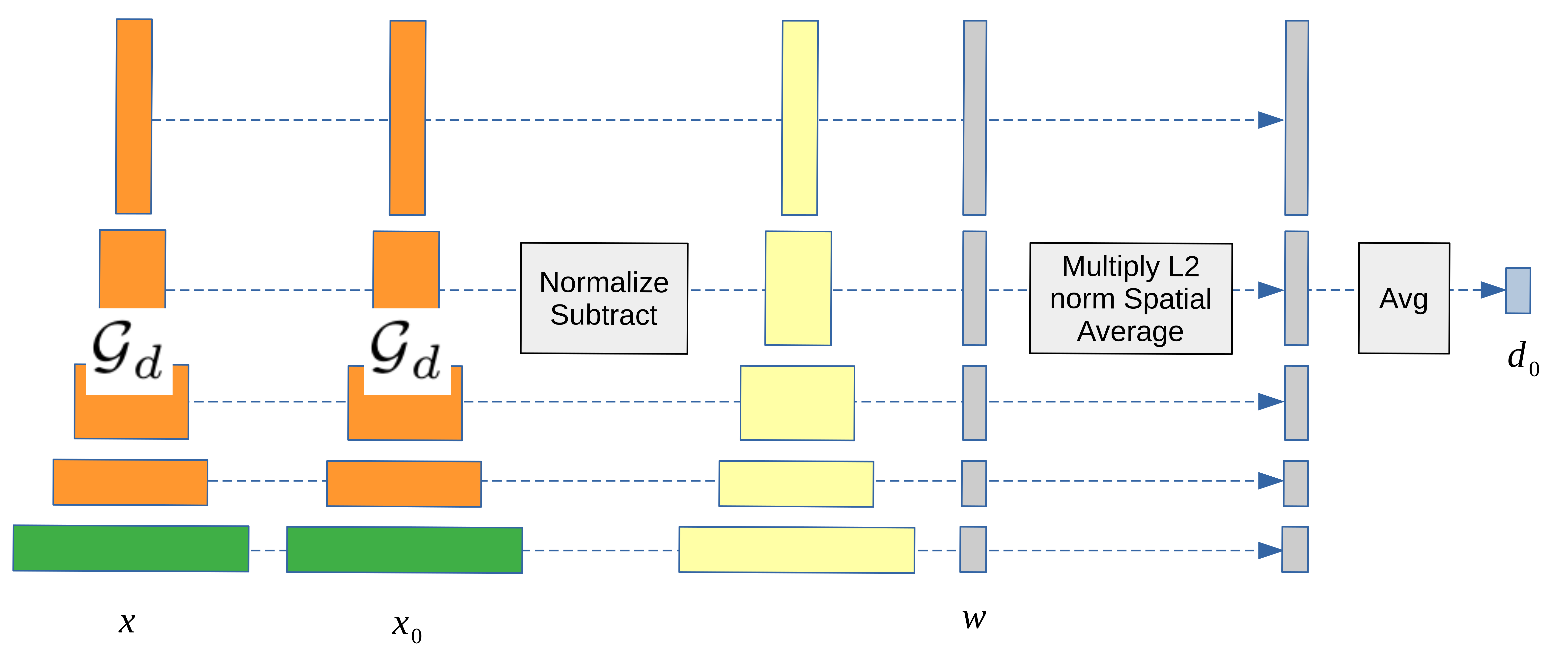}
\caption{Computing Euclidean distances from feature outputs $e^l$ and $\hat{y}^5$ between images $x$ and $x_0$. }
\label{fig:cpips}
\end{figure}

To obtain the distance between two images, denoted as $x$ and $x_0$, we follow the same procedure as LPIPS \cite{zhang2018perceptual} by learning a linear layer $w$ on the BAPPS dataset. This linear layer assigns weights to the compressed latents and intermediate decoder outputs. Fig. \ref{fig:cpips} illustrates the process of obtaining the distance using entropy-decoded $\hat{y}^5$ and feature outputs $e^l$ from our decoder network $\mathcal{G}_d$. We extract feature maps $\hat{y}^5,e^l \in \mathbb{R}^{C_l\times H_l\times W_l}$ for all layers $l$ and normalize them in the channel dimension. The activations are then scaled channel-wise using the vector $w^l \in \mathbb{R}^{C_l}$, and the $l_2$ distance is computed. Finally, we average across the spatial dimensions and all layers to obtain the following:

\begin{equation}
d(f^l) = \frac{1}{H_lW_l} \sum_{h,w} || w_l \odot (f_{hw}^l - f_{0hw}^l) ||_2^2 
\end{equation}

Eq. \eqref{eq_1} calculates the final distance between image $x$ and $x_0$, that is:

\begin{equation}
d_0 = \sum_{l=1}^4 d(e^l) + d(\hat{y}^5)
\label{eq_1}
\end{equation}

Furthermore, we train another smaller network, denoted as $\mathcal{D}$, to predict perceptual judgments $h$ from the distance pair $(d_0,d_1)$ on the BAPPS 151k patches 2AFC (two alternative forced choice) dataset.

\section{Experimental Results}

\subsection{Experimental Settings}

To implement our CPIPS, we utilize the CompressAI\footnote{https://github.com/InterDigitalInc/CompressAI} \cite{begaint2020compressai} implementation of the hyperprior neural compressor \cite{balle2018variational} and the official release\footnote{https://github.com/richzhang/PerceptualSimilarity} of LPIPS. We pre-train the image classification task on the ImageNet dataset, which consists of 1.2 million images. The training is performed using the PyTorch Adam optimizer with a learning rate 0.0001 for 120 epochs. Following that, we jointly train the compression-classification task with the pre-trained weights for 150 epochs, employing the Adam optimizer with a learning rate of 0.0001.

Regarding CPIPS weights $w$ and the judgment network $\mathcal{D}$, we train them for ten epochs using the BAPPS 2AFC dataset, as mentioned in the original LPIPS paper.

\subsection{Left-UNet Image Classification}

Table \ref{tab:imagenet-acc} displays our top-1 and top-5 accuracy compared to high-capacity deep networks such as VGG-16. The achieved top-1 accuracy of 60.11\% is considered favorable, indicating that the pre-trained weights can serve as a suitable initialization for the Left-UNet encoder $\mathcal{G}_e$.

\begin{table}[!ht]
\caption{ImageNet classification accuracy}
\label{tab:imagenet-acc}
\centering
\begin{tabular}{lrr}
\hline
\textbf{Network} & \textbf{Top-1 Acc.} & \textbf{Top-5 Acc.} \\ \hline
AlexNet & 56.52\% & 79.06\% \\
Left-UNet & 60.11\% & 81.95\% \\
ResNet18 & 69.36\% & 89.03\% \\
VGG-16 & 71.51\% & 93.38\%  \\ \hline
\end{tabular}
\end{table}

\subsection{Human Judgment Accuracy}

\begin{figure}[!t]
\centering
\includegraphics[width=1\columnwidth]{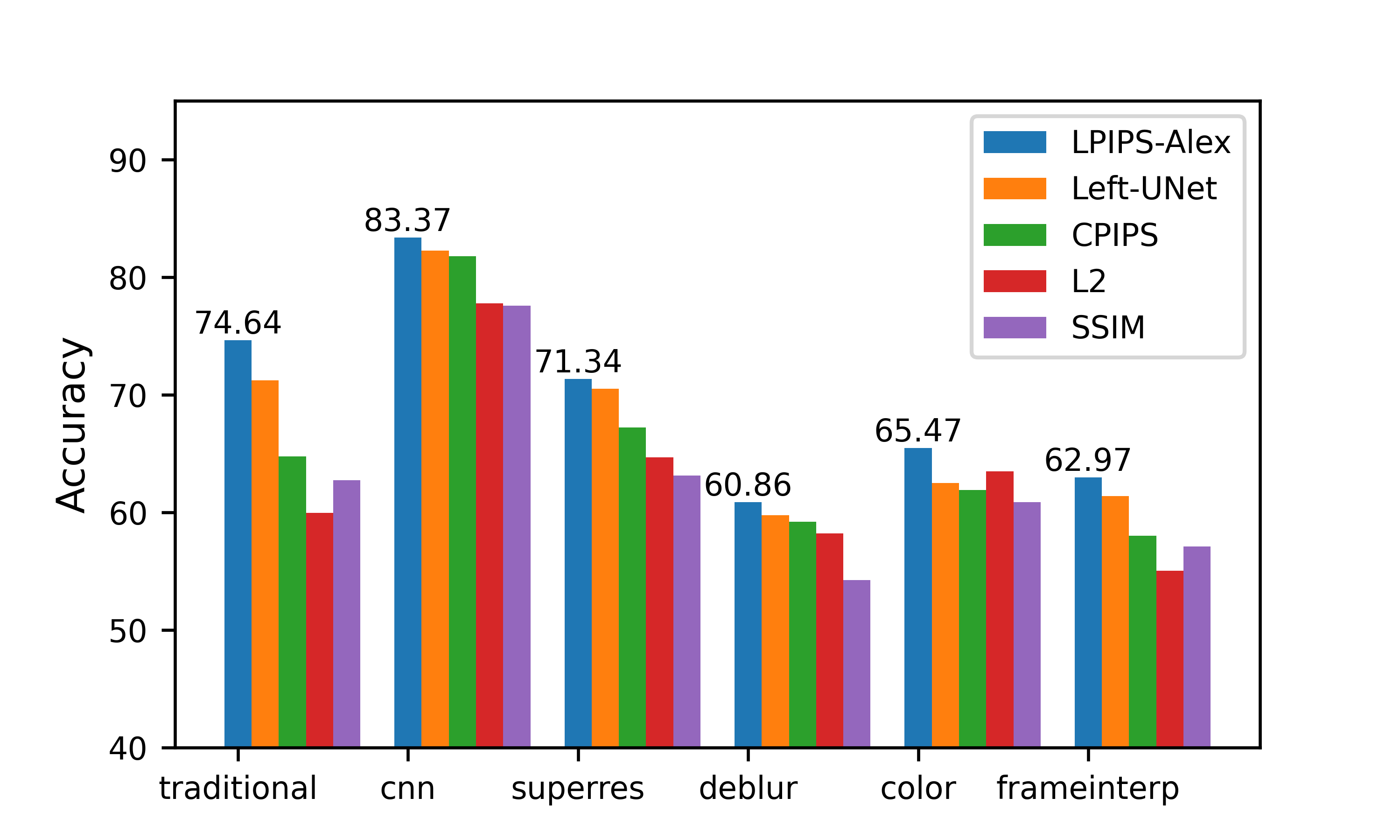}
\caption{Comparison of 2AFC accuracy against human ratings on the BAPPS dataset.}
\label{fig:bapps-2afc}
\end{figure}

We compare our method with LPIPS and  traditional L2 and SSIM metrics, in terms of the accuracy of image judgments against human ratings on the BAPPS dataset. Table \ref{tab:bapps-2afc} and Fig. \ref{fig:bapps-2afc} present the results.

\begin{table}[!ht]
\centering
\caption{2AFC judgment accuracy}
\begin{tabular}{lrrrrrr}
\hline
Method & Trad. & CNN & S.Res & DeBlur & Color & F.Interp \\ \hline
LPIPS-Alex  & 74.64 & 83.37 & 71.34 & 60.86 & 65.47 & 62.97 \\
Left-UNet & 71.23  & 82.27  & 70.51 & 59.74 & 62.50 & 61.39 \\
CPIPS & 64.77 & 81.77 & 67.21 & 59.20 & 61.91 & 58.00 \\
L2 & 59.94 & 77.76 & 64.67 & 58.19 & 63.50 & 55.02 \\
SSIM & 62.73 & 77.59 & 63.13 & 54.23 & 60.88 & 57.10 \\
\hline
\end{tabular} \label{tab:bapps-2afc}
\end{table}

Evidently, the metrics incorporating learned semantic features, such as LPIPS, Left-UNet, and CPIPS, exhibit a higher correlation with human judgments compared to L2 and SSIM. While Left-UNet does not achieve the same level of accuracy as LPIPS, it serves as an upper bound for our proposed CPIPS since they share the same feature extraction convolution layers. Our CPIPS achieves similar accuracy to Left-UNet in the CNN, DeBlur, and Color subsets but experiences a more considerable drop in accuracy in the Traditional, Super-Res, and Frame-Interp subsets. We attribute this drop to two factors: 1) the rate-distortion optimization process influencing the semantic properties of the latent vectors, thereby affecting the perceptual representation, and 2) the multi-scale feature maps $e^l$ serving as proxies for the feature extraction vectors $y^l$ reconstructed in the decoding stages through the regularization loss. Investigating and improving upon these factors are left as future work.

\begin{figure*}[!t]
\centering
\includegraphics[width=2.1\columnwidth]{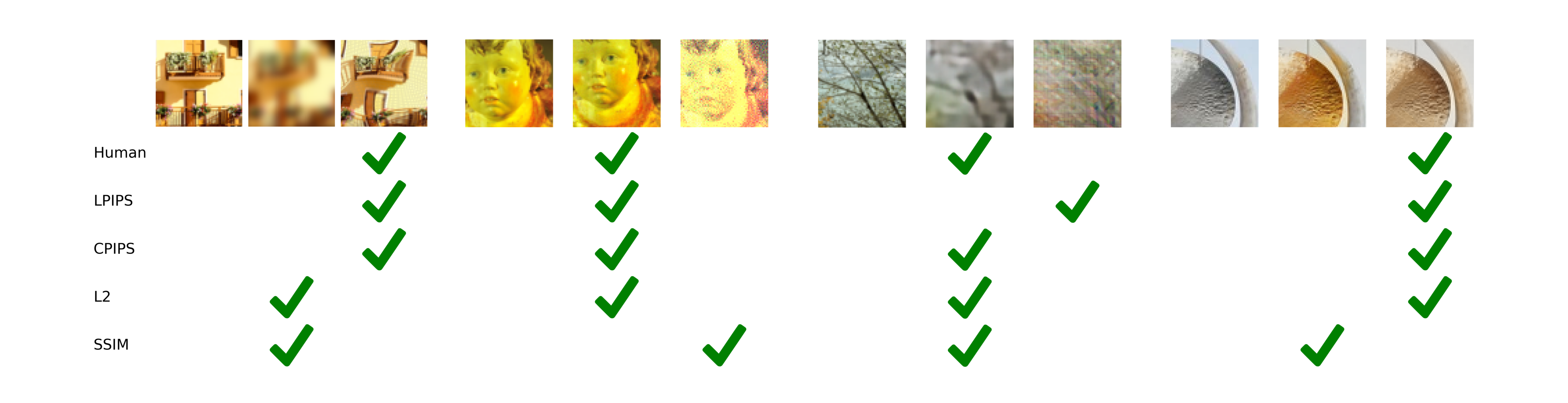}
\caption{The qualitative comparison of selected samples from the BAPPS dataset. The image sets are from the Traditional subset for the first two, CNN and Color subset for the third and last. The orders in each image set are the reference image, the distorted patch-0, and the distorted patch-1.}
\label{fig:bapps-quality}
\end{figure*}

Qualitatively, we select some sample image patches from the BAPPS dataset and present their different judgments in Fig. \ref{fig:bapps-quality}. We can see that the L2 and SSIM cannot reflect human perceptual preferences. At the same time, the CPIPS and LPIPS align with the ground truth better. The second image pair in Fig. \ref{fig:bapps-quality} demonstrates that the SSIM has a strong bias with structures and tends to be impacted by additive noises.

\subsection{Computational Complexity}

We assessed the computation time of the metrics on an Intel i7-9700K workstation with an Nvidia GTX 3090 GPU. To compare our CPIPS metric with LPIPS and DISTS\footnote{https://github.com/dingkeyan93/DISTS}, we used the Kodak dataset \cite{kodakcd} and calculated the average time cost, as shown in Table \ref{tab:perf}. Due to utilizing of a less complex neural network that only requires decoding the bitstream and intermediate features, our CPIPS method is approximately 50 times faster.

\begin{table}[!ht]
\centering
\caption{Metric Computation Time on Kodak}
\begin{tabular}{lrr}
\hline
Method & Avg. Time (secs.) \\ \hline
CPIPS & \textbf{0.0205} \\
LPIPS-Alex & 1.0681 \\
DISTS & 1.0373 \\
\hline
\end{tabular} \label{tab:perf}
\end{table}




\section{Conclusions}

In this work, we have introduced an end-to-end learned approach for image compression that aims to preserve perceptual distances. By leveraging pre-training on an image classification task and joint compression-classification training, we initialize the parameters of a learned image coding model with semantic features and guide the gradient descent process to emphasize semantic relevance. We have proposed a UNet-inspired network architecture Left-UNet, shared between the image classifier and the image encoder. Our approach calculates the difference in feature vectors between rate-distortion optimized compressed latents and intermediate decode outputs of two images, providing a perceptual distance preserving metric. We refer to this metric as CPIPS, derived from a learned image codec bitstream at no additional cost. Our experimental results demonstrate that CPIPS aligns more with human subjective judgments than traditional distortion metrics such as L2 and SSIM.

\section*{Acknowledgment}
The authors would like to thank the NSTC of Taiwan and CITI SINICA for supporting this research under the grant numbers 111-2221-E-002-134-MY3 and Sinica 3012-C3447.


\printbibliography

\end{document}